\documentclass[10pt,twocolumn,letterpaper]{article}

\usepackage{wacv}              

\makeatletter
\renewcommand\paragraph{\@startsection{paragraph}{4}{\z@}%
  {1ex plus .2ex minus .1ex}
  {-0.6em}
  {\normalfont\normalsize\bfseries}}
\makeatother

\usepackage{stfloats}

\definecolor{wacvblue}{rgb}{0.21,0.49,0.74}
\usepackage[breaklinks,colorlinks,allcolors=wacvblue]{hyperref}
\usepackage{multirow}

\def\wacvPaperID{61} 
\def\confName{WACV}
\def\confYear{2027}

\title{COGS: A Diagnostic Benchmark for Attribute-Aware Object-Level 3D Gaussian Splatting Perception}

\author{%
Zheng Xiao$^{1}$ \quad
Yifan Liu$^{2}$ \quad
Weijian Deng$^{3}$ \quad
Jiashu Li$^{2}$\\
Ziqin Liu$^{2}$ \quad
Zhenjun Han$^{2}$\thanks{Corresponding author.} \quad
Jianbin Jiao$^{2}$\\[3pt]
$^{1}$School of Advanced Interdisciplinary Sciences,
University of Chinese Academy of Sciences, China\\
$^{2}$University of Chinese Academy of Sciences, China\\
$^{3}$Tsinghua Shenzhen International Graduate School, China\\[3pt]
{\tt\small xiaozheng19@mails.ucas.ac.cn,
hanzhj@ucas.ac.cn}
}

\begin{document}
\maketitle
\begin{abstract}
3D Gaussian Splatting (3DGS) is increasingly explored as an input representation for 3D perception. Yet it remains unclear how current 3D architectures exploit heterogeneous Gaussian attributes. This ambiguity makes accuracy hard to interpret and limits principled architecture design for 3DGS perception. However, existing object-level 3DGS datasets lack the appearance variation or artifact control needed for systematic attribute-level analysis.
We introduce \textbf{COGS}, a large-scale 3DGS dataset with 15,675 artifact-filtered and appearance-rich objects across 40 categories. Built on COGS, we establish \textbf{COGS-Bench} for object classification and attribute-usage diagnosis, complemented by external validation. Our benchmark shows that Gaussian attributes are not uniformly beneficial: selected subsets can outperform the full attribute set, and the effects of appearance, opacity, and covariance-related attribute combinations vary substantially across architectures. We further study this finding with \textbf{HALO}, a reference model that preserves attribute-specific evidence before hierarchical aggregation. Experiments on COGS-Bench and external 3DGS datasets demonstrate that COGS supports systematic diagnosis of 3DGS attribute usage, while HALO highlights the value of role-preserving attribute organization. Our dataset is available at
\url{https://huggingface.co/datasets/XiaoZhengCV/COGS}.
\end{abstract}
    
\section{Introduction}

\begin{figure}[t]
    \centering
    \includegraphics[width=\linewidth]{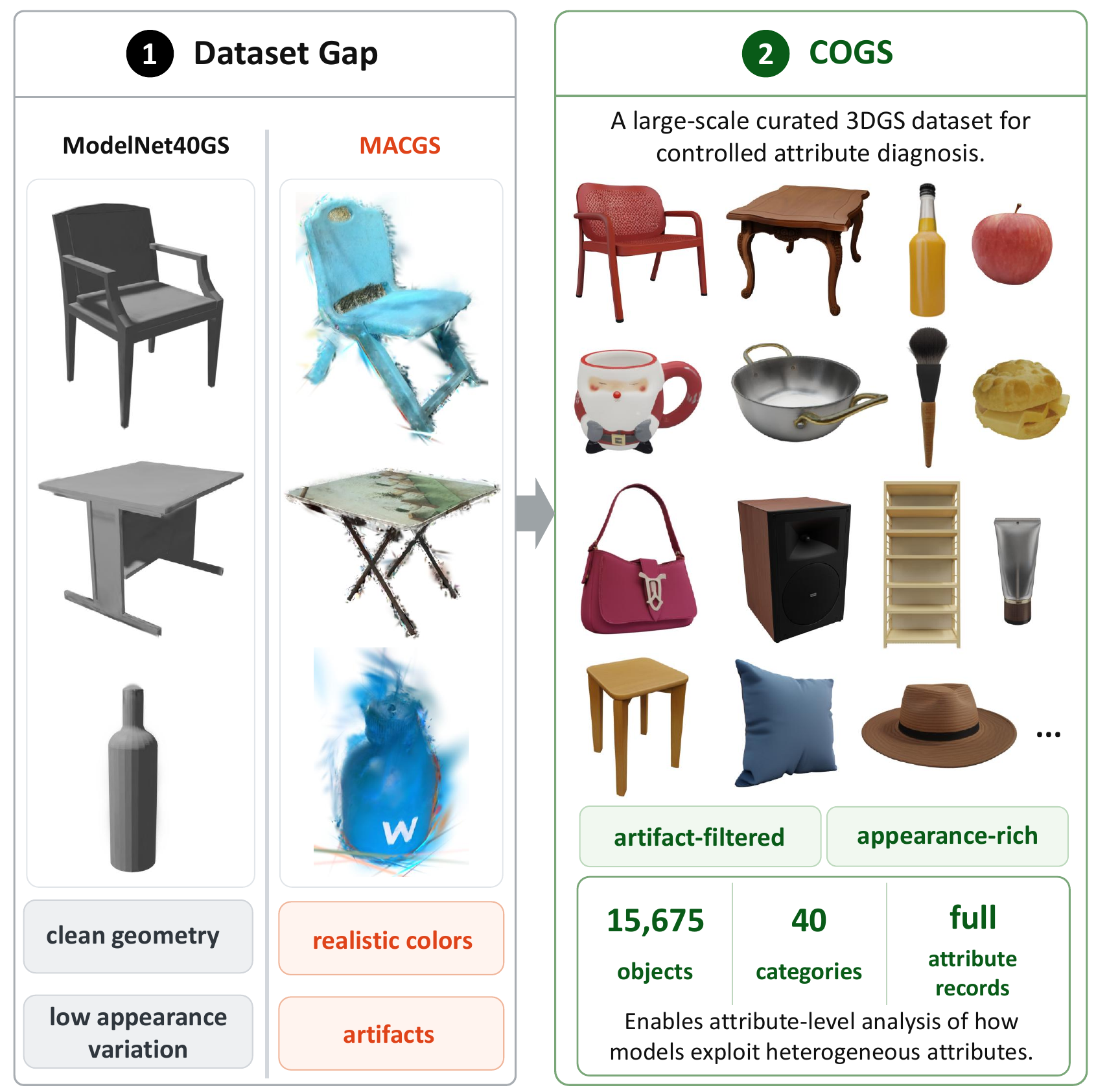} 
    \caption{Overview of COGS and COGS-Bench. Existing object-level 3DGS datasets provide complementary but limited properties, motivating COGS as a curated, artifact-filtered, and appearance-rich dataset for controlled attribute diagnosis. 
    }
    \label{fig:dataset_samples_compare}
\end{figure}

3D Gaussian Splatting (3DGS) has become a high-fidelity and real-time 3D representation for novel-view synthesis~\cite{kerbl20233d}. Beyond rendering, recent studies increasingly explore 3DGS for 3D perception. Some methods~\cite{guo2026semantic,zhou2024feature,wu2024opengaussian,qin2024langsplat,ye2024gaussian,cen2025segment,choi2024click} use Gaussian primitives as carriers of distilled 2D semantic or language features~\cite{cen2025segment,radford2021learning,oquab2023dinov2} through multi-view feature lifting and language-field optimization. More recent works directly process Gaussian primitives for object recognition and scene understanding~\cite{ma2024shapesplat,zhang2025mitigating,li2025scenesplat}. These studies demonstrate the potential of 3DGS for perception, but they also expose a representation-level ambiguity: it remains unclear whether current 3D architectures truly exploit Gaussian-specific attributes, or instead obtain their gains from positional or appearance shortcuts.

This ambiguity is important because 3DGS primitives are not ordinary point samples with extra feature channels. Each primitive is parameterized by position, opacity, scale, rotation, and appearance coefficients. These attributes are jointly optimized for rendering, but they encode different aspects. Position specifies spatial support, scale and rotation describe anisotropic extent, opacity is related to visibility, and appearance coefficients provide color-related cues. Therefore, the benefit of 3DGS for perception should not be viewed simply as adding more input channels to point coordinates. Treating these heterogeneous attributes as homogeneous point features may underuse informative cues or introduce interference. A systematic benchmark is therefore needed to disentangle architecture design and attribute usage under unified protocols.

Scene-level 3DGS datasets are essential for evaluating scalability and contextual reasoning~\cite{li2025scenesplat}, but recognition in scenes is entangled with background context, occlusion, layout, and co-occurrence, making them less suitable for controlled attribute diagnosis. Object-level datasets provide a complementary setting, yet existing datasets remain limited as shown in \cref{fig:dataset_samples_compare}. CAD-derived datasets such as ModelNet40GS~\cite{ma2024shapesplat} provide clean geometry but limited appearance variation, while capture-based datasets such as MACGS~\cite{zhang2025mitigating} contain realistic observations but may include reconstruction artifacts. A diagnostic benchmark for 3DGS perception should therefore provide object-level controllability, rich appearance cues, artifact-filtered assets, and full Gaussian attribute records.

To this end, we introduce \textbf{COGS}, \textbf{C}ommon \textbf{O}bjects of 3D \textbf{G}aussian \textbf{S}platting, a large-scale object-centric 3DGS dataset for controlled analysis of Gaussian attribute utilization. COGS contains 15,675 objects across 40 categories and is constructed from diverse object images through a curated generative pipeline with multi-stage filtering and manual 3DGS inspection. The resulting dataset provides rich appearance variation, artifact-filtered structures, and full Gaussian attribute records for controlled attribute-level diagnosis. 
Built on COGS, we establish \textbf{COGS-Bench} for object classification and attribute-usage diagnosis, evaluating recognition accuracy, efficiency, and attribute combinations under consistent protocols, with additional validation on other 3DGS datasets. As an additional study, we examine COGS as a pre-training source for object classification on MACGS.
Beyond diagnosis, COGS-Bench also enables us to examine an attribute-aware modeling principle: models should preserve attribute-specific evidence from Gaussian attributes with different representational roles before high-level semantic aggregation. We instantiate this principle with \textbf{HALO}, Hierarchical Aggregation with Locally-decOupled MLP, as a reference model. HALO separately extracts local evidence from Gaussian attribute groups and fuses them through hierarchical aggregation. HALO provides a concrete example of translating attribute-usage
diagnosis into architecture design for 3DGS perception.

Our contributions are summarized as follows:
\begin{itemize}
\item We introduce \textbf{COGS}, a large-scale 3DGS dataset containing 15,675 objects across 40 categories, with artifact-filtered structures and rich appearance variation.
\item We establish \textbf{COGS-Bench} for object classification and attribute-usage diagnosis under consistent protocols, with cross-pipeline validation on external 3DGS datasets.
\item We provide systematic evidence that Gaussian attributes are not uniformly beneficial across architectures and that more attributes do not always lead to better recognition.
\item We provide \textbf{HALO}, an attribute-aware reference model that preserves attribute-specific evidence before hierarchical semantic aggregation, illustrating how
attribute-usage diagnosis can inform model design.
\end{itemize}
   
\section{Related Work}

\subsection{Semantic 3DGS via 2D Feature Lifting}

Recent studies have extended 3DGS beyond photorealistic rendering by associating semantic or language features with Gaussian primitives. Feature 3DGS~\cite{zhou2024feature} learns semantic feature fields on top of 3DGS, enabling semantic segmentation and language-guided editing. LangSplat~\cite{qin2024langsplat} distills and compresses CLIP-related language features into 3D Gaussians, constructing a compact 3D language field for efficient open-vocabulary localization. Gaussian Grouping~\cite{ye2024gaussian} introduces an identity encoding for each Gaussian and supervises it with multi-view 2D masks and 3D consistency constraints, enabling object grouping, 3D segmentation, and local editing. Segment Any 3D Gaussians~\cite{cen2025segment} and Click-Gaussian~\cite{choi2024click} further leverage 2D segmentation masks or interactive cues for 3DGS segmentation. 

\subsection{Direct 3DGS Perception and Benchmarks}

Beyond using 3DGS as a carrier for distilled 2D features, recent works have begun to treat Gaussian primitives themselves as inputs for 3D perception. MACGS~\cite{zhang2025mitigating} studies object classification with 3DGS. It constructs an object-level 3DGS dataset from real-world captures in MVImgNet~\cite{yu2023mvimgnet} and evaluates multiple point-cloud classification architectures by comparing position-only inputs with non-color Gaussian attributes. ShapeSplat~\cite{ma2024shapesplat} builds object-level 3DGS datasets from commonly used 3D asset datasets~\cite{chang2015shapenet,yi2016scalable}. It further introduces Gaussian-MAE, which adapts masked autoencoding from point clouds to Gaussian primitives for self-supervised representation learning. SceneSplat~\cite{li2025scenesplat} extends direct Gaussian primitive processing to scene-level understanding. It constructs 3DGS scenes from indoor datasets~\cite{dai2017scannet,yeshwanth2023scannet++,straub2019replica,roberts2021hypersim,wald2019rio,baruch2021arkitscenes,chang2017matterport3d} and adopts a PTv3-style backbone~\cite{wu2024point} with self-supervised and vision-language pretraining, supporting both closed-set semantic segmentation and open-vocabulary scene understanding.



\section{COGS and COGS-Bench}
\label{sec:cogs}

\begin{figure*}[t] 
\centering 
\includegraphics[width=\linewidth]{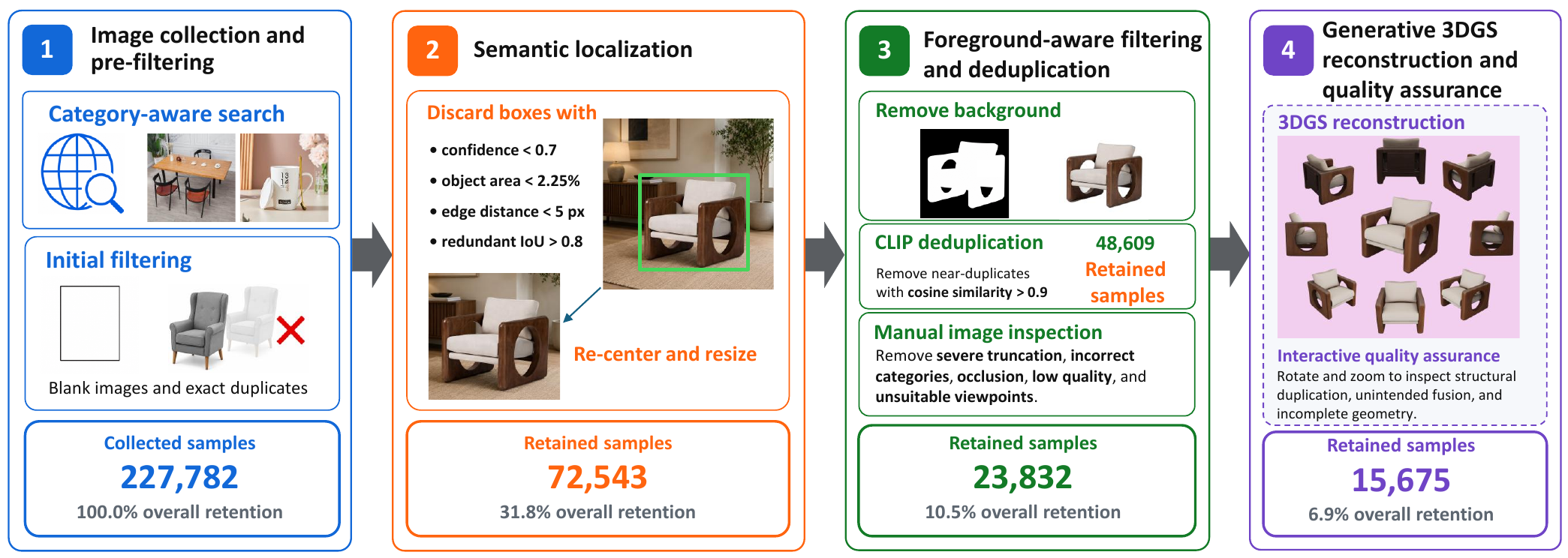} 
\caption{Overview of the COGS construction pipeline. Starting from large-scale web image collection, we perform semantic localization, foreground-aware filtering and deduplication, manual image inspection, generative 3DGS reconstruction with TRELLIS, and viewer-based 3DGS quality inspection to obtain curated object-centric 3DGS assets.} 
\label{fig:cogs_pipeline} 
\end{figure*}

\begin{figure}[t]
    \centering
    \includegraphics[width=\linewidth]{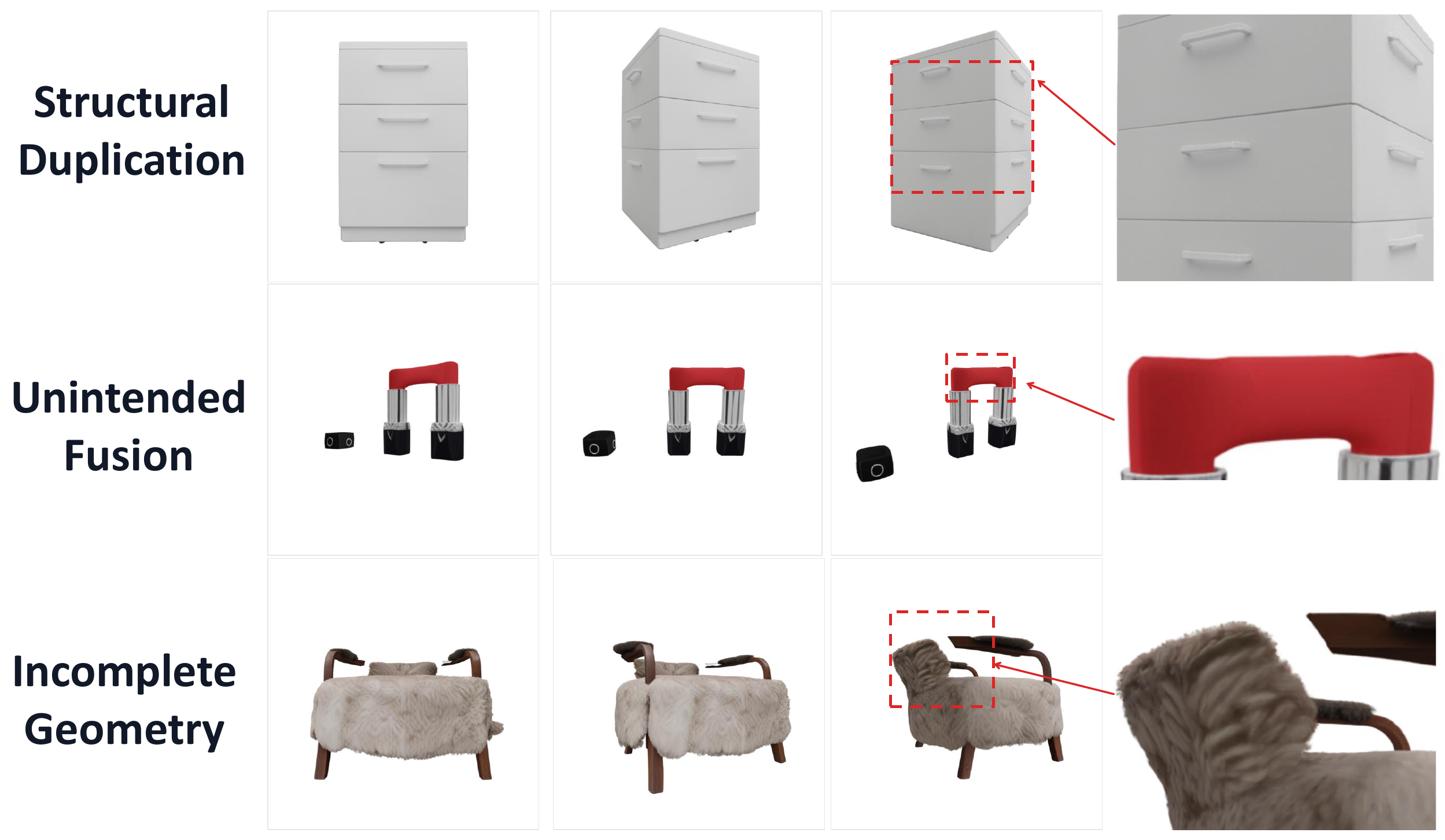} 
    \caption{Discarded 3DGS samples rendered from multiple views. Row 1 shows view-inconsistent structural duplication, where drawer features expected on one face also appear on adjacent faces; row 2 shows unintended fusion between two lipstick bodies; row 3 shows incomplete chair geometry.}
    \label{fig:dataset_samples_fail}
\end{figure}

\subsection{Dataset Construction}
\label{subsec:cogs_construction}

Capture-derived object assets can suffer from unobserved contact regions and mask leakage, producing incomplete geometry or floating artifacts. COGS therefore uses the scalable generative pipeline in \cref{fig:cogs_pipeline} as a complementary object-centric testbed with reduced capture noise, rich appearance, and full Gaussian attributes, rather than as a replacement for captured datasets.

\paragraph{Image collection and pre-filtering.}
We construct COGS from 227,782 raw object images discovered through established image-search services using category-specific queries. During acquisition, our automated data-collection pipeline checks license fields exposed in the search results together with the associated source-page metadata. We retain only images marked as CC0 or Public Domain. Samples with missing, conflicting, or unverifiable license information, NoDerivatives restrictions, or other terms incompatible with reuse, modification, and redistribution are discarded. Generic category names and fine-grained descriptors of material, style, shape, and subtype improve intra-class coverage; their overlapping results also motivate the subsequent deduplication. Blank images and exact duplicates are removed first.

\paragraph{Semantic localization.}
Web images often contain multiple objects, cluttered backgrounds, or irrelevant content. To obtain object-centric inputs, we localize the target object using text-prompted object detection with GroundingDINO~\cite{liu2024grounding}. Detections with confidence $<0.7$, object area $<2.25\%$, edge distance $<5$ pixels, or redundant overlap (IoU $>0.8$) are discarded; the remaining boxes are cropped with a small margin and re-centered.

\paragraph{Foreground-aware filtering and deduplication.} 
U2-Net foreground masks~\cite{qin2020u2} reduce background bias in similarity estimation. For incomplete or occluded foregrounds, diffusion inpainting~\cite{Labs2025FLUX1KF} provides more complete appearances for CLIP feature extraction. These inpainted images are used only for CLIP-based deduplication. Samples with cosine similarity $>0.9$ are deduplicated~\cite{radford2021learning}. Manual image inspection then removes severe truncation, incorrect or ambiguous categories, heavy occlusion, low visual quality, and unsuitable viewpoints.

\paragraph{Generative 3DGS reconstruction and quality assurance.}
Each retained image is converted into a 3DGS asset using TRELLIS~\cite{xiang2025structured}. 
Three annotators independently inspect each sample in the SuperSplat Editor by rotating and zooming it through arbitrary viewpoints. As illustrated in \cref{fig:dataset_samples_fail}, we flag (i) view-inconsistent structural duplication, where a feature expected on one surface is spuriously repeated on adjacent surfaces; (ii) unintended fusion between parts that should remain separate; and (iii) incomplete geometry with missing major structures. Only assets unanimously judged acceptable are retained; this conservative quality-control stage reduces 23,832 candidates to 15,675 assets.

\begin{figure}[t]
    \centering
    \includegraphics[width=\linewidth]{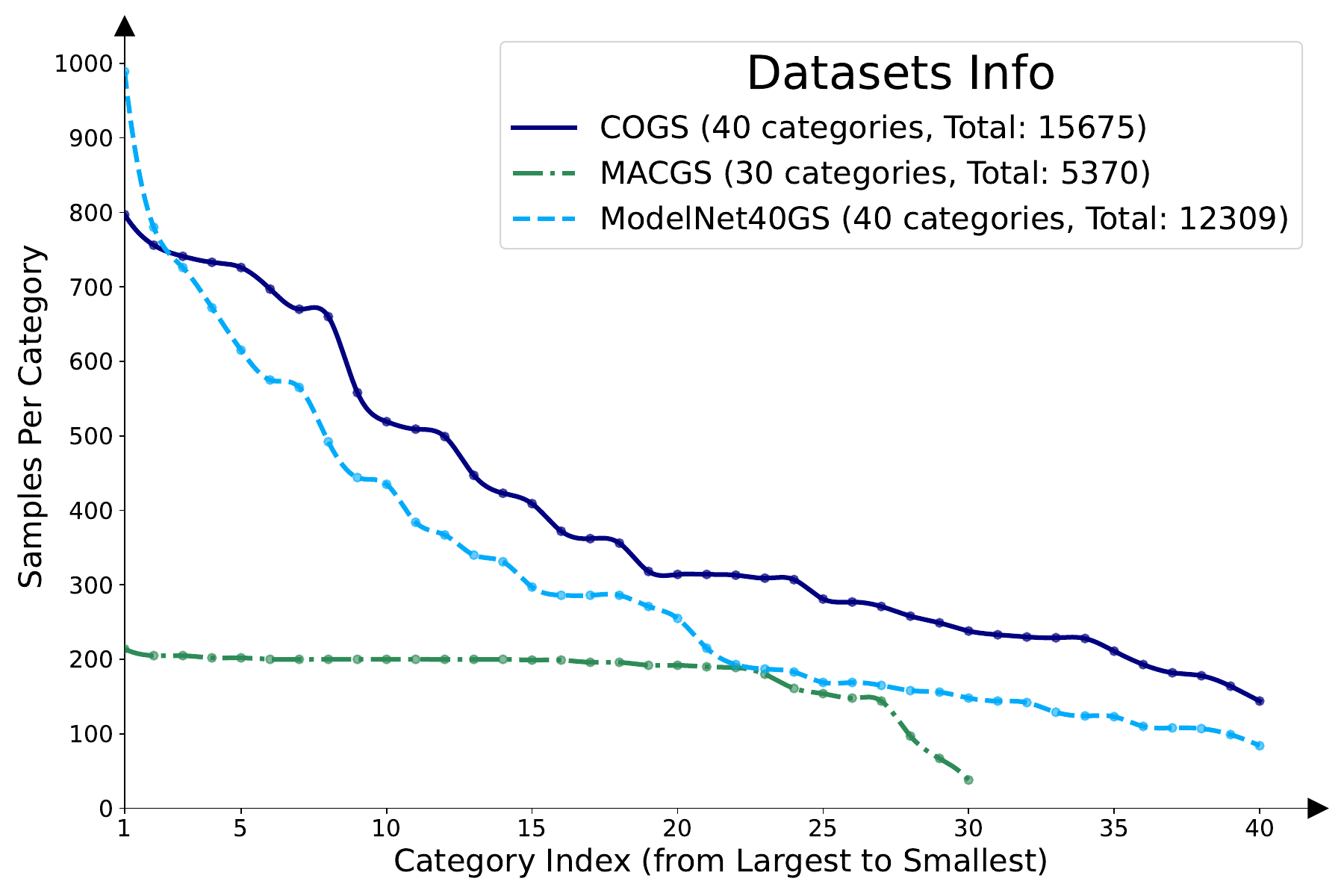} 
    \caption{Category distribution comparison. Categories in each dataset are independently sorted by object count. The x-axis represents dataset-specific sorted category indices rather than aligned category identities. MACGS statistics use the authors' publicly released 30-category dataset.}
    \label{fig:cogs_distribution}
\end{figure}

Overall, the pipeline reduces 227,782 images to 15,675 curated 3DGS objects across 40 categories, prioritizing object validity, diversity, and artifact control over scale. \Cref{fig:cogs_distribution} shows the distribution; category counts and additional visualizations are provided in the supplement.

\subsection{COGS-Bench Protocol}
\label{subsec:cogs_bench}
Built on COGS, we establish COGS-Bench for object classification and attribute-usage diagnosis. The dataset is released for non-commercial research and educational purposes. Rights-holder removal requests will be supported.

\paragraph{Object classification.}
The primary task is 40-class classification, evaluated by overall accuracy (OA) and mean class accuracy (mAcc). Following the common train/test-only benchmark protocol in point-cloud classification~\cite{qi2017pointnet,wang2019dynamic}, we use no separate validation split. To control the long-tailed distribution, we use a fixed training set of 100 instances per category (4,000 in total), with the remaining 11,675 instances reserved for testing.

\paragraph{Attribute usage diagnosis.}
COGS-Bench evaluates eight input configurations: \textbf{P}, \textbf{PO}, \textbf{PC}, \textbf{POC}, \textbf{PSQ}, \textbf{POSQ}, \textbf{PSQC}, and \textbf{POSQC}, where \textbf{P}, \textbf{O}, \textbf{S}, \textbf{Q}, and \textbf{C} denote position, opacity, scale, quaternion/rotation, and appearance/color. Position is retained as the common structural reference because the evaluated backbones use it for neighborhood construction and sampling; removing it would change the backbone rather than only its input attributes. Since scale and rotation jointly parameterize anisotropic covariance, \textbf{SQ} is treated as one functional group. The eight settings therefore cover all combinations of the optional groups \textbf{O}, \textbf{C}, and \textbf{SQ}, with \textbf{P} always included. This protocol compares how attribute combinations affect recognition across the evaluated architectures.

\paragraph{External validation.}
We repeat the eight-configuration attribute diagnosis on MACGS~\cite{zhang2025mitigating} and further examine attribute usage and attribute-specific processing in 3DGS semantic segmentation on ScanNet~\cite{li2025scenesplat}. These experiments assess whether the diagnostic findings extend to capture-derived objects and real-world scenes.

\section{HALO: Attribute-aware Reference Model}
\label{sec:method}

\begin{figure*}[t]
\centering
\includegraphics[width=0.9\textwidth]{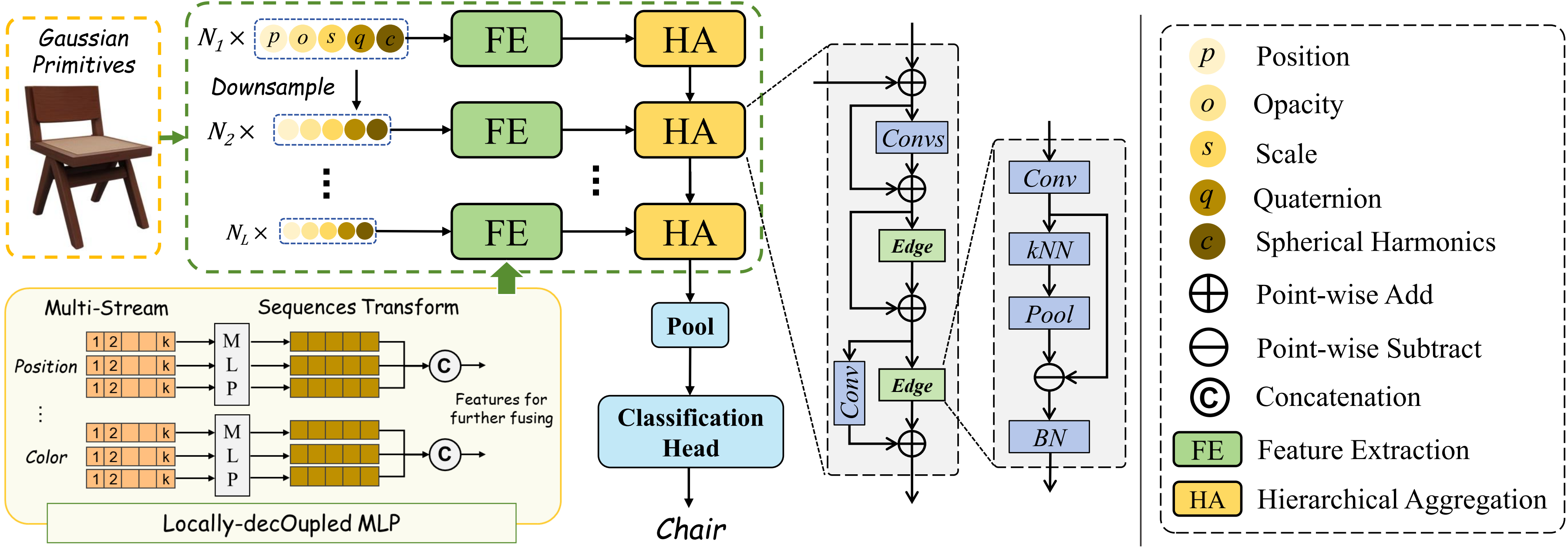}
\caption{Overview of HALO. Given a set of Gaussian primitives, HALO progressively builds hierarchical representations through repeated Feature Extraction (FE), Hierarchical Aggregation (HA), and downsampling. FE performs multi-stream attribute encoding with Locally-decOupled MLP (LOM), while HA aggregates neighborhood evidence before primitive decimation.}
\label{fig:halo_overview}
\end{figure*}

\subsection{Motivation and Overview}
\label{subsec:halo_overview}

HALO is designed to examine an attribute-aware modeling principle suggested by the heterogeneous nature of 3DGS primitives: models should preserve attribute-specific evidence associated with attributes that play different representational roles. Directly treating them as a single descriptor may force one encoder to learn attribute-specific patterns and cross-attribute interactions simultaneously.
Based on this principle, we introduce \textbf{HALO}, Hierarchical Aggregation with
Locally-decOupled MLP, as an attribute-aware reference model for COGS-Bench. 
HALO first extracts local evidence within attribute-specific streams and then fuses the resulting descriptors through hierarchical aggregation. As shown in \cref{fig:halo_overview}, each hierarchy level contains a Feature Extraction (FE) module and a Hierarchical Aggregation (HA) module. FE combines attribute-specific local descriptors with point-wise features, while HA aggregates neighborhood context before downsampling. The final representation is globally pooled for object-level recognition.

\subsection{Attribute-aware Feature Extraction}
\label{subsec:attribute_aware_fe}

The FE module is designed to avoid premature homogenization of Gaussian attributes. Let $\mathcal{A}=\{p,o,s,q,c\}$ denote the attribute groups, corresponding to position, opacity, scale, rotation, and color, and let $\mathbf{z}^a_i\in\mathbb{R}^{d_a}$ be the value of attribute group $a$ for primitive $i$. For attribute-specific encoding in the decoupled FE module, we convert normalized quaternions to axis-angle vectors and DC spherical-harmonic coefficients to RGB, which serve as the rotation and color inputs, respectively. For each primitive, HALO first extracts local descriptors in attribute-specific streams and then fuses the resulting features. 

For primitive $i$, we construct a $k$-nearest-neighbor set $\mathcal{N}_k(i)$ in coordinate space. For the $m$-th component of attribute group $a$, we form a compact neighbor sequence
\begin{equation}
\Delta \mathbf{z}^{a}_{i,m}
=[z^{a}_{j_1,m}-z^{a}_{i,m},\ldots,z^{a}_{j_k,m}-z^{a}_{i,m}],
j_r\in\mathcal{N}_k(i),
\label{eq:lom_sequence}
\end{equation}
where neighbors are ordered by Euclidean distance to obtain a fixed-length sequence. The Locally-decOupled MLP (LOM) projects each component-wise sequence and concatenates the component descriptors:
\begin{equation}
\mathbf{l}^{a}_{i}
=\operatorname{Cat}_{m=1}^{d_a}
\rho_a(\Delta \mathbf{z}^{a}_{i,m}),
\label{eq:lom_descriptor}
\end{equation}
where $\rho_a$ is an attribute-specific sequence projection. Unlike per-neighbor local MLPs that repeatedly process $N\times k$ neighbor features, LOM treats each component-wise neighborhood as a compact sequence. The neighbor dimension is mainly handled in the first projection layer, making multi-stream attribute modeling more efficient.

In parallel, HALO encodes point-wise attribute values with two mappings rather than one mapping for each attribute. Specifically, we keep position as an individual point-wise stream and concatenate all non-position attributes into a second stream:
\begin{equation}
\begin{aligned}
\mathbf{u}^{p}_{i} = \eta_p(\mathbf{p}_i), \quad
\mathbf{u}^{np}_{i} = \eta_{np}(\operatorname{Cat}(o_i,\mathbf{s}_i,\mathbf{q}_i,\mathbf{c}_i)),
\end{aligned}
\label{eq:pointwise_descriptor}
\end{equation}
where $\eta_p$ and $\eta_{np}$ are point-wise MLPs for the position and non-position streams, respectively. 
The FE output is obtained by fusing local and point-wise descriptors in the position and non-position streams:
\begin{equation}
\begin{aligned}
F^{\mathrm{FE}}_i
&=\phi_p\big(\operatorname{Cat}(\mathbf{l}^{p}_{i},\mathbf{u}^{p}_{i})\big) 
+\phi_{np}\big(\operatorname{Cat}(\mathbf{l}^{np}_{i},\mathbf{u}^{np}_{i})\big),
\end{aligned}
\label{eq:fe_fusion}
\end{equation}
where $\phi_p$ and $\phi_{np}$ are projection functions, and $\mathbf{l}^{np}_{i}=\operatorname{Cat}(\mathbf{l}^{o}_{i},\mathbf{l}^{s}_{i},\mathbf{l}^{q}_{i},\mathbf{l}^{c}_{i})$. This formulation delays cross-attribute interaction until after attribute-specific local modeling.

\subsection{Hierarchical Aggregation}
\label{subsec:hierarchical_aggregation}

HALO progressively reduces the number of primitives to build an efficient hierarchical encoder. However, directly downsampling Gaussian primitives may remove local evidence that is useful for recognition. The HA module therefore aggregates neighborhood context before each decimation step.

At hierarchy level $l$, let $F_l^{\mathrm{FE}}$ be the output of FE. For $l>1$, features inherited from the previous level are aligned to the current sampled primitives and fused with current features:
\begin{equation}
\begin{aligned}
F_l^{\mathrm{fuse}} = U_l + \theta_l(U_l), \quad U_l = F_l^{\mathrm{FE}} + D_l(F_{l-1}),
\end{aligned}
\label{eq:cross_level_fusion}
\end{equation}
where $D_l(\cdot)$ denotes index-based downsampling with channel projection, and $\theta_l$ is a residual~\cite{he2016deep} mapping. For the first level, $F_1^{\mathrm{fuse}}=F_1^{\mathrm{FE}}$.

Given a feature set $H=\{\mathbf{h}_i\}$, an edge-pooling block~\cite{li2019deepgcns}  retains crucial local information within each neighborhood:
\begin{equation}
\begin{aligned}
\Omega_l(H)_i
= \max_{j\in\mathcal{N}_k(i)}
\big(\widehat{\mathbf{h}}_j-\widehat{\mathbf{h}}_i\big), \quad \widehat{\mathbf{h}}_i = \omega_l(\mathbf{h}_i),
\end{aligned}
\label{eq:edge_pooling}
\end{equation}
where $\omega_l$ is a linear projection and the maximum is taken channel-wise over neighbors. Two residual edge-pooling blocks are stacked before downsampling:
\begin{equation}
\begin{aligned}
\widetilde{F}_l
&= F_l^{\mathrm{fuse}}+\Omega_{l,1}(F_l^{\mathrm{fuse}}), \\
F_l
&= \gamma_l(\widetilde{F}_l)+\Omega_{l,2}(\widetilde{F}_l),
\end{aligned}
\label{eq:ha_stack}
\end{equation}
where $\gamma_l$ aligns feature dimensions when necessary. The resulting features are passed to the next hierarchy level after downsampling. In this way, HALO follows an aggregation-before-decimation strategy: local evidence around retained primitives is first absorbed into their features, and primitive density is then reduced for efficient high-level reasoning.

\section{Experiments and Diagnostic Analysis}
\label{sec:experiments}

Our experiments cover dataset quality and characterization, classification and attribute usage, HALO component ablations, and external validation. We additionally examine COGS-to-MACGS transfer.

\subsection{Experimental Setup}
\label{subsec:exp_setup}
For classification, we sample 1,024 primitives from each 3DGS object using FPS~\cite{qi2017pointnet++}. Before FPS, COGS objects undergo random rotations followed by removal of a spatially contiguous region containing 5--20\% of the primitives around a randomly selected primitive. Across datasets, we activate opacity and scale using sigmoid and exponential functions, respectively, center positions and normalize positions and scales by the object radius, normalize quaternions, and retain only DC appearance coefficients. All classification experiments use random seed 42. Unless otherwise specified, supervised models are trained from random initialization under the same protocol. We train each model for 200 epochs using AdamW~\cite{loshchilov2017decoupled} with a batch size of 32 on a single RTX 4090 GPU. The learning rate is initialized to $1\times10^{-3}$ and decayed to $1\times10^{-5}$ using a cosine annealing scheduler. Gaussian-MAE~\cite{ma2024shapesplat} follows its masked pre-training protocol. For HALO, we evaluate two variants, HALO-B as the recommended accuracy--efficiency configuration and HALO-L as a capacity-scaling variant. Both use a three-level hierarchy containing 1,024, 256, and 64 primitives, with 20 neighbors at every level. HALO-B uses feature dimensions $[64,128,256]$, while HALO-L widens them to $[256,512,1024]$.

\begin{table}[t]
\centering
\caption{Rendered visual quality comparison. We report MUSIQ and CLIPIQA+ scores on rendered 48-view and single-view images. Standard deviations are shown in parentheses.}
\label{tab:dataset_comparison}
\resizebox{\linewidth}{!}{
\begin{tabular}{l|cc|cc}
\toprule
\multirow{2}{*}{Dataset}
& \multicolumn{2}{c|}{48 views} & \multicolumn{2}{c}{Single view} \\
\cmidrule(lr){2-3} \cmidrule(lr){4-5}
& MUSIQ ($\uparrow$) & CLIPIQA+ ($\uparrow$) & MUSIQ ($\uparrow$) & CLIPIQA+ ($\uparrow$) \\
\midrule
ModelNet40GS~\cite{ma2024shapesplat}
& $58.71\,(\pm 3.39)$ & $0.5228\,(\pm 0.0158)$
& $52.25\,(\pm 5.33)$ & $0.4969\,(\pm 0.0162)$ \\
MACGS~\cite{zhang2025mitigating}
& $38.64\,(\pm 3.87)$ & $0.4509\,(\pm 0.0256)$
& $49.94\,(\pm 5.23)$ & $0.5172\,(\pm 0.0233)$ \\
\textbf{COGS}
& $\mathbf{64.75\,(\pm 3.84)}$ & $\mathbf{0.6013\,(\pm 0.0278)}$
& $\mathbf{63.21\,(\pm 4.54)}$ & $\mathbf{0.5863\,(\pm 0.0343)}$ \\
\bottomrule
\end{tabular}
}
\end{table}

\subsection{Dataset Quality and Characterization}
\label{subsec:dataset_quality}

\paragraph{Rendered visual quality.}
We assess 3DGS asset quality using rendered RGB views, where severe geometry or appearance defects become directly observable. To reduce bias from the long-tailed category distributions, stratified sampling selects up to 80 objects per category with seed 42, yielding 3,200 COGS, 3,200 ModelNet40GS, and 2,345 MACGS objects. Scores are macro-averaged over categories. For each selected instance, we render 48 views on four spherical elevation rings ($0^{\circ}$, $15^{\circ}$, $30^{\circ}$, and $45^{\circ}$), with 12 uniformly spaced azimuths per ring. Adjacent rings are offset by $15^{\circ}$ in azimuth to improve coverage. The single-view setting uses the canonical front view at $0^{\circ}$ elevation and $0^{\circ}$ azimuth. We use two no-reference image quality metrics: MUSIQ~\cite{ke2021musiq} for perceptual quality and technical distortions, and CLIPIQA+~\cite{wang2023exploring} for high-level perceptual realism.
As shown in \cref{tab:dataset_comparison}, COGS achieves the best results in both single-view and 48-view settings. MACGS~\cite{zhang2025mitigating} obtains a competitive single-view CLIPIQA+ score, but its 48-view scores drop substantially, suggesting that artifacts and incomplete geometry become more visible under broader viewpoint coverage. In contrast, COGS maintains strong scores across both settings. Together with the manual 3DGS inspection in \cref{subsec:cogs_construction}, these results support COGS as a curated object-level benchmark for controlled 3DGS attribute diagnosis.

\begin{figure}[t]
\centering
\includegraphics[width=\linewidth]{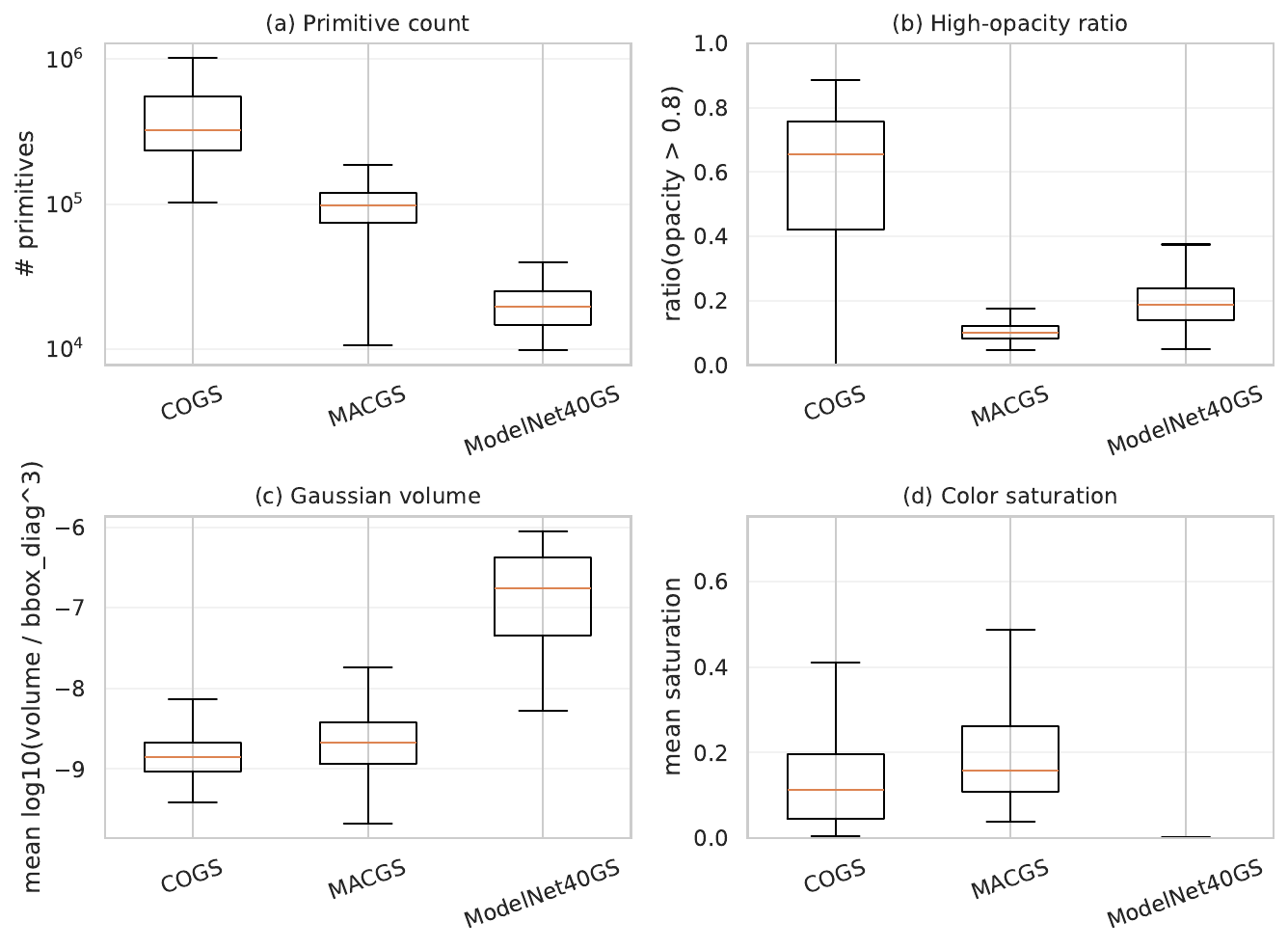}
\caption{Gaussian attribute statistics across COGS, MACGS, and ModelNet40GS. We compare primitive count, high-opacity ratio ($>$0.8), normalized Gaussian volume, and color saturation.}
\label{fig:attr_statistics}
\end{figure}

\paragraph{Gaussian attribute statistics.}
We next compare Gaussian attribute statistics: the number of Gaussian primitives, the fraction of primitives with activated opacity $>0.8$, the mean log-normalized Gaussian volume, and the mean color saturation. For a primitive with scales $(s_x,s_y,s_z)$ and object bounding-box diagonal $d_{\mathrm{bbox}}$, the normalized volume statistic is computed as $\log_{10}(s_xs_ys_z/d_{\mathrm{bbox}}^3)$ and then averaged over primitives. For color, we convert the degree-0 spherical harmonic coefficients to RGB, compute the channel range $\max(R,G,B)-\min(R,G,B)$ for each primitive, and average it over the object. \cref{fig:attr_statistics} summarizes these distributions using boxplots. 
The three datasets exhibit markedly different distributions. COGS contains more primitives and a higher ratio of high-opacity Gaussians. ModelNet40GS has fewer primitives and near-zero color saturation because it is derived from CAD models without rich texture information. MACGS lies between them in primitive count, but shows lower high-opacity ratios and more concentrated color statistics. These dataset-specific attribute distributions motivate the external validation in \cref{subsec:external_validation}.

\begin{figure}[t]
\centering
\includegraphics[width=0.9\linewidth]{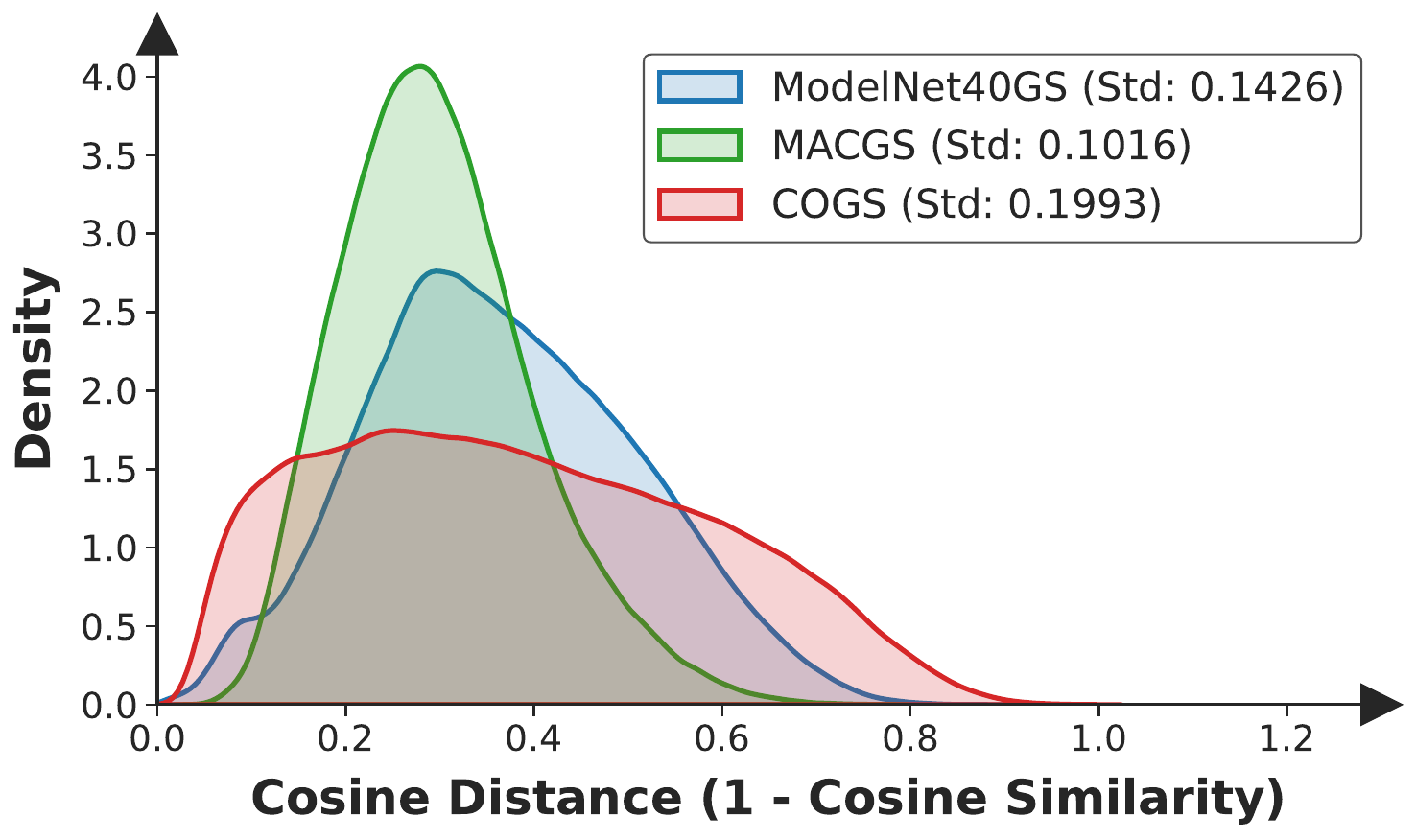}
\caption{Intra-class pairwise cosine distance distribution in the DINOv2 feature space. COGS shows the largest spread, indicating richer intra-class diversity than ModelNet40GS and MACGS.}
\label{fig:diversity_histogram_arrow}
\end{figure}

\paragraph{Intra-class diversity.}
We further measure intra-class diversity in the rendered image feature space. For each object, we render four views, extract DINOv2 features~\cite{oquab2023dinov2}, and average the view-level features into an object-level representation. We then compute intra-class pairwise cosine distances. \cref{fig:diversity_histogram_arrow} shows the kernel density estimation of the resulting distance distributions. COGS has the largest spread, with a standard deviation of 0.1993, compared with 0.1426 for ModelNet40GS and 0.1016 for MACGS. This suggests that COGS contains richer rendered visual intra-class variation. 

\subsection{COGS-Bench}
\label{subsec:exp_cogs_bench}

\paragraph{Classification benchmark.}
\cref{tab:cogs_results} reports classification results on COGS using all Gaussian attributes. PTv3~\cite{wu2024point} and Gaussian-MAE~\cite{ma2024shapesplat} reach 89.11\% and 88.20\% OA, while the attribute-aware HALO-B/L references obtain 94.50\%/94.70\%. These results show that COGS-Bench is non-trivial for existing architectures and provides a standardized testbed for studying 3DGS attribute modeling. Supplementary Table~\ref{tab:supp_inference_efficiency} further reports counted FLOPs, measured inference latency and throughput, and peak inference memory. 

\begin{table}[t]
\centering
\caption{Classification benchmark on COGS with all Gaussian attributes as input.}
\label{tab:cogs_results}
\setlength{\tabcolsep}{4pt}
\resizebox{\linewidth}{!}{
\begin{tabular}{l|ccccc}
\toprule
Method & OA (\% $\uparrow$) & mAcc (\% $\uparrow$) & Mem. (MB $\downarrow$)
& Time (min $\downarrow$) & Params (M $\downarrow$) \\
\midrule
PointNet~\cite{qi2017pointnet}
& 79.53 & 79.70 & 1112 & \textbf{17.40} & 3.47 \\
DGCNN~\cite{wang2019dynamic}
& 85.69 & 86.70 & 5411 & 61.09 & \textbf{1.81} \\
PTv1~\cite{zhao2021point}
& 87.67 & 88.32 & \textbf{744} & 72.80 & 3.26 \\
DeLA~\cite{chen2023decoupled}
& 88.48 & 89.14 & 1771 & \underline{33.06} & 5.34 \\
PTv3~\cite{wu2024point}
& 89.11 & 89.77 & 5710 & 212.15 & 38.87 \\
Gaussian-MAE~\cite{ma2024shapesplat}
& 88.20 & 88.58 & 2876 & 49.31 & 22.10 \\
\midrule
\textbf{HALO-B (Ours, Ref.)}
& \underline{94.50} & \underline{94.70} & \underline{750} & 39.60 & \underline{2.68} \\
\textbf{HALO-L (Ours, Ref.)}
& \textbf{94.70} & \textbf{94.91} & 1786 & 41.60 & 10.46 \\
\bottomrule
\end{tabular}
}
\end{table}

\begin{table*}[ht]
\centering
\caption{Attribute usage diagnosis on COGS-Bench. P, O, S, Q, and C denote position, opacity, scale, rotation, and appearance/color, respectively. Each entry reports OA/mAcc (\%).}
\label{tab:cogs_ablation_attr}
\scriptsize
\setlength{\tabcolsep}{6pt}
\resizebox{\linewidth}{!}{
\begin{tabular}{l|cccccccc}
\toprule
Method & P & PO & PC & POC & PSQ & POSQ & PSQC & POSQC \\
\midrule
PointNet~\cite{qi2017pointnet}
& 73.25/68.48 & 80.73/79.84 & 76.52/75.82 & 77.27/77.24
& 81.23/79.31 & 82.71/81.30 & 78.93/78.49 & 79.53/79.70 \\
DGCNN~\cite{wang2019dynamic}
& 85.87/86.15 & 87.04/87.31 & 82.79/83.84 & 84.17/85.20
& 88.57/88.42 & 89.18/89.36 & 85.29/86.08 & 85.69/86.70 \\
PTv1~\cite{zhao2021point}
& 86.45/86.43 & 87.10/87.65 & 87.85/88.34 & 88.09/88.76
& 88.54/88.27 & 88.82/89.26 & 88.46/89.15 & 87.67/88.32 \\
DeLA~\cite{chen2023decoupled}
& 89.28/89.97 & 91.21/91.68 & 83.17/84.31 & 85.45/86.54
& 91.17/91.33 & 90.40/91.00 & 87.58/88.49 & 88.48/89.14 \\
PTv3~\cite{wu2024point}
& 88.53/88.80 & 89.50/89.89 & 89.16/89.16 & 88.58/88.89 & 88.74/89.01 & 89.68/89.82 & 88.44/88.68 & 89.11/89.77 \\
Gaussian-MAE~\cite{ma2024shapesplat}
& 87.01/86.32 & 89.18/89.49 & 85.89/85.79 & 86.56/87.43
& 91.71/91.09 & 89.10/88.96 & 87.01/87.18 & 88.20/88.58 \\
\midrule
\textbf{HALO-B (Ours, Ref.)}
& 89.66/89.88 & 91.31/91.85 & 91.82/92.63 & 92.35/93.09
& 93.34/93.07 & 93.70/93.95 & 93.93/94.00 & 94.50/94.70 \\
\textbf{HALO-L (Ours, Ref.)}
& 90.56/91.07 & 92.23/92.82 & 92.57/93.11 & 93.13/94.02
& 93.88/94.12 & 94.30/94.40 & 94.54/94.94 & 94.70/94.91 \\
\bottomrule
\end{tabular}
}
\end{table*}

\begin{table}[t]
\centering
\caption{Ablation on attribute-specific processing in HALO-B.}
\label{tab:attr_ablation_HALO}
\setlength{\tabcolsep}{6pt}
\resizebox{\linewidth}{!}{
\begin{tabular}{lcc}
\toprule
HALO-B & OA (\% $\uparrow$) & mAcc (\% $\uparrow$) \\
\midrule
Shared attribute processing & 89.03 & 89.67 \\
Attribute-specific processing & 94.50 & 94.70 \\
\bottomrule
\end{tabular}
}
\end{table}

\paragraph{Attribute usage diagnosis.}
Beyond final classification accuracy, COGS-Bench enables a controlled diagnosis of how architectures use heterogeneous Gaussian attributes. We denote position, opacity, scale, quaternion/rotation and appearance/color as P, O, S, Q, and C, respectively. Each entry in \cref{tab:cogs_ablation_attr} reports OA/mAcc (\%) under a specific attribute subset.
The results show that adding more Gaussian attributes does not uniformly improve recognition. For early-concatenation backbones, selected attribute subsets can outperform the full POSQC input. For example, PointNet~\cite{qi2017pointnet} and DGCNN~\cite{wang2019dynamic} perform best with POSQ rather than POSQC. 
These results indicate that simply adding more Gaussian attributes does not guarantee better recognition, and that heterogeneous attributes may introduce optimization or fusion conflicts when they are collapsed into a single feature space too early.
COGS-Bench therefore diagnoses architecture-dependent attribute utilization rather than prescribing a universal best subset. HALO-B/L are reference responses to this diagnosis: after attribute-specific encoding and delayed fusion, both attain their best OA with full POSQC.
HALO-B improves from 89.66\%/89.88\% with position only to 91.31\%/91.85\% with PO, 93.70\%/93.95\% with POSQ, and 94.50\%/94.70\% with POSQC. HALO-L follows the same trend. 
Overall, these results highlight the main purpose of COGS-Bench: it is not merely an accuracy leaderboard, but a diagnostic benchmark for studying when and how Gaussian attributes contribute to object-level 3DGS perception.

\paragraph{Ablation on attribute-specific processing.}
We ablate HALO-B by replacing its attribute-specific feature extraction with a shared stream over concatenated Gaussian attributes. As shown in \cref{tab:attr_ablation_HALO}, shared processing reduces performance from 94.50\%/94.70\% to 89.03\%/89.67\% in OA/mAcc, confirming that preserving attribute-specific evidence before fusion is critical for HALO.

\subsection{External Validation}
\label{subsec:external_validation}
\paragraph{Object-level cross-pipeline validation.}
We repeat the attribute diagnosis on capture-derived MACGS. With full results in Supplementary~\cref{tab:supp_macgs_attribute}, \cref{tab:macgs_diagnostic_summary} summarizes the complete eight-configuration diagnosis. All six conventional backbones attain higher OA with at least one attribute subset than with POSQC, with gains of 0.28--3.05 points. In contrast, HALO-B/L attain their best OA with full POSQC, reaching 93.99\%/94.55\%. The COGS diagnosis therefore persists on a capture-derived distribution with substantially different Gaussian statistics. In addition, Supplementary~\cref{tab:supp_external_results} reports the full-POSQC comparison on capture-derived MACGS and CAD-derived ModelNet40GS.

\begin{table}[t]
\centering
\caption{Cross-pipeline attribute diagnosis on MACGS (OA).}
\label{tab:macgs_diagnostic_summary}
\setlength{\tabcolsep}{4pt}
\resizebox{\linewidth}{!}{
\begin{tabular}{lc}
\toprule
Models & Diagnostic result (OA, \%) \\
\midrule
6 conventional baselines & 6/6: subset $>$ POSQC (+0.28--3.05) \\
HALO-B / HALO-L & POSQC best: 93.99 / 94.55 \\
\bottomrule
\end{tabular}
}
\end{table}

\paragraph{Real-world scene-level validation.}
We further test subset-over-full behavior and the benefit of attribute-specific processing in 3DGS semantic segmentation on ScanNet~\cite{li2025scenesplat} using the official 10\% Limited-Reconstruction split~\cite{hou2021exploring}. As summarized in \cref{tab:scannet_diagnostic_summary} and detailed in Supplementary~\cref{tab:supp_scannet_semseg}, the complete PTv1 diagnosis shows that PO, POSQ, and PSQC all outperform POSQC in mIoU, with POSQ improving from 48.45\% to 50.37\% (+1.92). This is composition-dependent rather than a general preference for fewer attributes, since the other subsets perform below POSQC. Under the identical full POSQC input, attribute-specific HALO improves shared processing from 43.81\% to 49.83\% mIoU (+6.02). These results extend both diagnostic findings from COGS to captured objects and real-world scenes.

\begin{table}[t]
\centering
\caption{Scene-level validation on ScanNet.}
\label{tab:scannet_diagnostic_summary}
\setlength{\tabcolsep}{4pt}
\resizebox{\linewidth}{!}{
\begin{tabular}{llcc}
\toprule
Method & Variant/Input & mIoU (\%) & OA (\%) \\
\midrule
PTv1 & POSQC & 48.45 & 80.95 \\
     & POSQ  & \textbf{50.37} & 81.08 \\
HALO & Shared POSQC & 43.81 & 79.16 \\
     & Attribute-specific POSQC & \textbf{49.83} & \textbf{81.96} \\
\bottomrule
\end{tabular}
}
\end{table}

\subsection{Transfer and Low-Shot Adaptation}
\label{subsec:transfer}
As an additional study, we examine COGS as a pre-training source for object classification on MACGS. For COGS pre-training, we use a fixed train/test split that assigns 80\% of the objects in each category to training and the remaining 20\% to testing. Each model is then fully fine-tuned using 10\%, 20\%, 50\%, or 100\% of the MACGS training instances in each class. Scratch and pre-trained runs differ only in initialization.
We consider representative backbone families, including point-wise MLPs (PointNet~\cite{qi2017pointnet}), graph-based local aggregation (DGCNN~\cite{wang2019dynamic}), transformer-based attention (PTv1~\cite{zhao2021point}), and HALO-B as the attribute-aware reference model.
Despite the distribution gap in \cref{fig:attr_statistics}, \cref{tab:cross_results} shows consistent gains from COGS pre-training. With only 10\% MACGS training data, OA improves by 26.98, 21.91, 12.01, and 3.23 points for PointNet, DGCNN, PTv1, and HALO-B, respectively. The gains decrease as more target data becomes available but remain positive for every evaluated architecture. Full scratch and pre-trained OA/mAcc results are provided in Supplementary~\cref{tab:supp_transfer_full}.

\begin{table}[t]
\centering
\caption{Models are fine-tuned on varying fractions of MACGS using either random initialization or COGS pre-training. Each cell reports pre-trained OA and OA gain over training from scratch. Full results are provided in the supplementary material.}
\label{tab:cross_results}
\setlength{\tabcolsep}{4pt}
\resizebox{\linewidth}{!}{
\begin{tabular}{l|cccc}
\toprule
Method & 10\% & 20\% & 50\% & 100\% \\
\midrule
PointNet~\cite{qi2017pointnet}
& 56.65 {\scriptsize(+26.98)} & 64.33 {\scriptsize(+18.58)} & 76.25 {\scriptsize(+7.03)} & 81.89 {\scriptsize(+2.13)} \\
DGCNN~\cite{wang2019dynamic}
& 60.54 {\scriptsize(+21.91)} & 71.35 {\scriptsize(+12.20)} & 81.79 {\scriptsize(+6.00)} & 88.82 {\scriptsize(+1.67)} \\
PTv1~\cite{zhao2021point}
& 62.29 {\scriptsize(+12.01)} & 73.29 {\scriptsize(+8.41)} & 84.94 {\scriptsize(+4.81)} & 90.11 {\scriptsize(+1.85)} \\
HALO-B
& 69.87 {\scriptsize(+3.23)} & 82.72 {\scriptsize(+2.04)} & 90.29 {\scriptsize(+1.10)} & 94.08 {\scriptsize(+0.09)} \\
\bottomrule
\end{tabular}
}
\end{table}

\section{Conclusion and Limitations}

We presented COGS, a large-scale object-centric 3DGS dataset with 15,675 artifact-filtered, appearance-rich assets across 40 categories, and COGS-Bench for object classification and attribute-usage diagnosis. Our experiments show that the usefulness of Gaussian attributes depends on the architecture and attribute combination, and that adding more attributes does not always improve recognition. HALO demonstrates the benefits of attribute-specific local encoding before fusion and hierarchical aggregation. External validation on MACGS and ScanNet further supports these diagnostic findings.

COGS focuses on controlled object-level diagnosis and does not replace scene-level or severity-stratified robustness benchmarks. Future work can extend COGS-Bench to controlled corruption tests, compositional scenes, dense annotations, and broader 3DGS construction pipelines.

\clearpage
\section*{Acknowledgements}
This work was supported in part by the Key Deployment Program of the Chinese Academy of Sciences, China under Grant KGFZD-145-25-39, the National Natural Science Foundation of China under Grants 62272438, and Beijing Natural Science Foundation L25700.
{
    \small
    \bibliographystyle{ieeenat_fullname}
    \bibliography{main}
}

\clearpage
\setcounter{page}{1}
\setcounter{section}{0}
\setcounter{table}{0}
\setcounter{figure}{0}
\renewcommand{\thepage}{S\arabic{page}}
\renewcommand{\thesection}{S\arabic{section}}
\renewcommand{\theHsection}{supp.\arabic{section}}
\renewcommand{\thetable}{S\arabic{table}}
\renewcommand{\thefigure}{S\arabic{figure}}

\twocolumn[{
\begin{center}
{\Large \textbf{Supplementary Material}}\\[0.5em]
\end{center}
}]

\section{Additional Dataset Details and Cross-Dataset Transfer}
\label{sec:supp_dataset_transfer}

\begin{figure*}[!b]
\centering
\includegraphics[width=\linewidth]{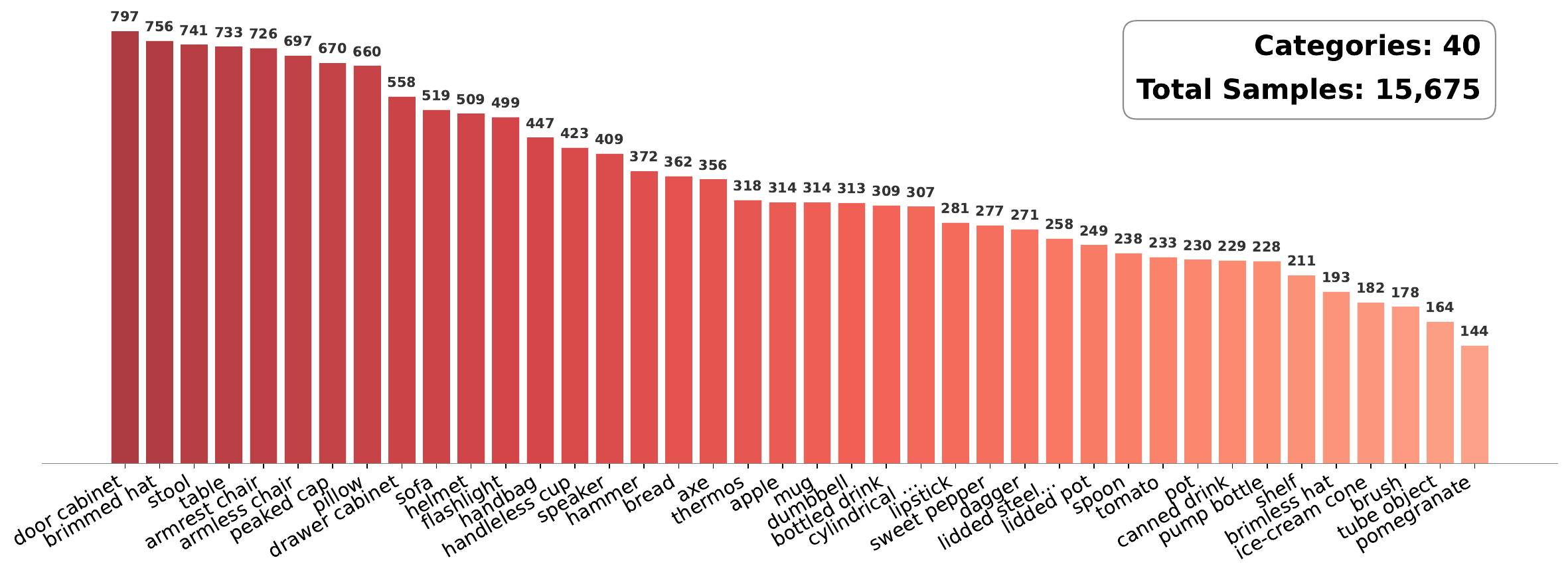}
\caption{
Category distribution of the COGS dataset. The figure reports the 40 category names and the corresponding number of curated 3DGS assets in each category. 
}
\label{fig:supp_detailed_cogs_distribution}
\end{figure*}

\paragraph{COGS Category Distribution.}
The detailed category names and per-category sample counts of COGS are illustrated in \cref{fig:supp_detailed_cogs_distribution}.

\paragraph{Full COGS-to-MACGS Transfer Results.}
We provide the full COGS-to-MACGS transfer results in \cref{tab:supp_transfer_full}. Compared with training from random initialization, COGS pre-training consistently improves all evaluated architectures across different target-data fractions. This result indicates that COGS provides transferable object-level 3DGS cues.

\begin{table}[t]
\centering
\caption{
Full cross-dataset transfer results from COGS to MACGS. Models are fine-tuned on varying fractions of MACGS using either random initialization or COGS pre-training. Gains over Scratch are shown in parentheses.
}
\label{tab:supp_transfer_full}
\setlength{\tabcolsep}{4pt}
\resizebox{\linewidth}{!}{
\begin{tabular}{lc|cc}
\toprule
\multirow{2}{*}{Method} & \multirow{2}{*}{Fraction}
& Scratch & Pre-trained on COGS \\
\cmidrule(lr){3-4}
& & OA / mAcc (\% $\uparrow$) & OA / mAcc (\% $\uparrow$) \\
\midrule
PointNet~\cite{qi2017pointnet} & \multirow{4}{*}{10\%}
& 29.67 / 29.12
& 56.65 {\scriptsize(+26.98)} / 54.51 {\scriptsize(+25.39)} \\
DGCNN~\cite{wang2019dynamic} &
& 38.63 / 37.28
& 60.54 {\scriptsize(+21.91)} / 58.42 {\scriptsize(+21.14)} \\
PTv1~\cite{zhao2021point} &
& 50.28 / 49.73
& 62.29 {\scriptsize(+12.01)} / 60.99 {\scriptsize(+11.26)} \\
HALO-B (Ours, Ref.) &
& 66.64 / 64.95
& 69.87 {\scriptsize(+3.23)} / 68.53 {\scriptsize(+3.58)} \\
\midrule
PointNet~\cite{qi2017pointnet} & \multirow{4}{*}{20\%}
& 45.75 / 45.01
& 64.33 {\scriptsize(+18.58)} / 62.84 {\scriptsize(+17.83)} \\
DGCNN~\cite{wang2019dynamic} &
& 59.15 / 57.50
& 71.35 {\scriptsize(+12.20)} / 69.28 {\scriptsize(+11.78)} \\
PTv1~\cite{zhao2021point} &
& 64.88 / 62.99
& 73.29 {\scriptsize(+8.41)} / 71.88 {\scriptsize(+8.89)} \\
HALO-B (Ours, Ref.) &
& 80.68 / 79.89
& 82.72 {\scriptsize(+2.04)} / 80.99 {\scriptsize(+1.10)} \\
\midrule
PointNet~\cite{qi2017pointnet} & \multirow{4}{*}{50\%}
& 69.22 / 68.75
& 76.25 {\scriptsize(+7.03)} / 74.99 {\scriptsize(+6.24)} \\
DGCNN~\cite{wang2019dynamic} &
& 75.79 / 74.38
& 81.79 {\scriptsize(+6.00)} / 80.09 {\scriptsize(+5.71)} \\
PTv1~\cite{zhao2021point} &
& 80.13 / 78.89
& 84.94 {\scriptsize(+4.81)} / 83.97 {\scriptsize(+5.08)} \\
HALO-B (Ours, Ref.) &
& 89.19 / 88.43
& 90.29 {\scriptsize(+1.10)} / 89.54 {\scriptsize(+1.11)} \\
\midrule
PointNet~\cite{qi2017pointnet} & \multirow{4}{*}{100\%}
& 79.76 / 77.97
& 81.89 {\scriptsize(+2.13)} / 80.80 {\scriptsize(+2.83)} \\
DGCNN~\cite{wang2019dynamic} &
& 87.15 / 85.80
& 88.82 {\scriptsize(+1.67)} / 87.70 {\scriptsize(+1.90)} \\
PTv1~\cite{zhao2021point} &
& 88.26 / 87.54
& 90.11 {\scriptsize(+1.85)} / 88.43 {\scriptsize(+0.89)} \\
HALO-B (Ours, Ref.) &
& 93.99 / 93.24
& 94.08 {\scriptsize(+0.09)} / 93.51 {\scriptsize(+0.27)} \\
\bottomrule
\end{tabular}
}
\end{table}

\section{Full Attribute Diagnosis on Capture-Derived MACGS}
\label{sec:supp_macgs_diagnosis}
To further examine whether the attribute-usage findings observed on COGS are specific to the TRELLIS generation pipeline, we additionally repeat the complete attribute diagnosis on MACGS. MACGS is constructed from real-world multi-view captures and exhibits substantially different Gaussian attribute statistics from COGS. We use the same eight input configurations as in the main COGS-Bench analysis: P, PO, PC, POC, PSQ, POSQ, PSQC, and POSQC. All models follow the same training and evaluation protocol used in the main paper. \cref{tab:supp_macgs_attribute} reports the complete results. The same attribute-usage behavior observed on COGS remains evident on MACGS. For every evaluated conventional backbone, at least one attribute subset outperforms the full POSQC input. For example, PointNet improves from 79.76\% OA with POSQC to 81.51\% with POC, DGCNN improves from 87.15\% to 90.20\%, and PTv1 improves from 88.26\% to 90.93\% with PSQC. Similar subset-over-full behavior is also observed for DeLA, PTv3, and Gaussian-MAE. In contrast, HALO-B and HALO-L achieve their best performance with the full POSQC input, reaching 93.99/93.24\% and 94.55/93.81\% OA/mAcc, respectively. These results provide an external diagnostic check beyond final classification accuracy: the observation that heterogeneous Gaussian attributes are not uniformly beneficial under conventional homogeneous processing, while HALO can more consistently exploit the complete attribute set, persists on a capture-derived 3DGS distribution. 

\begin{table*}[t] 
\centering 
\caption{ Complete attribute-usage diagnosis on MACGS. P, O, S, Q, and C denote position, opacity, scale, rotation, and appearance/color, respectively. Each entry reports OA/mAcc (\%). Bold numbers indicate the best OA for each method. } \label{tab:supp_macgs_attribute} 
\setlength{\tabcolsep}{3.2pt} 
\resizebox{\textwidth}{!}{ 
\begin{tabular}{lcccccccc} 
\toprule 
Method & P & PO & PC & POC & PSQ & POSQ & PSQC & POSQC \\ 
\midrule 
PointNet~\cite{qi2017pointnet} & 76.71/75.56 & 78.10/77.59 & 78.65/77.79 & \textbf{81.51}/80.83 & 77.36/76.24 & 76.06/75.65 & 78.16/77.61 & 79.76/77.97 \\ 
DGCNN~\cite{wang2019dynamic} & 88.91/88.24 & 87.89/86.71 & 88.17/87.28 & \textbf{90.20}/89.06 & 87.62/86.31 & 86.04/84.87 & 87.71/87.08 & 87.15/85.80 \\ 
PTv1~\cite{zhao2021point} & 88.91/87.79 & 87.80/86.68 & 87.99/87.33 & 88.82/88.46 & 89.00/88.11 & 89.37/88.85 & \textbf{90.93}/89.53 & 88.26/87.54 \\ 
DeLA~\cite{chen2023decoupled} & 91.50/90.43 & 91.50/90.47 & \textbf{92.51}/91.97 & 92.14/90.90 & 92.05/90.88 & 91.87/90.43 & 92.24/91.28 & 91.50/90.15 \\ 
PTv3~\cite{wu2024point} & 89.93/89.31 & 90.76/89.72 & 92.51/91.74 & 92.61/91.82 & 92.51/91.72 & \textbf{92.79}/92.17 & 92.61/92.03 & 92.33/91.45 \\ 
Gaussian-MAE~\cite{ma2024shapesplat} & 87.62/86.59 & 89.37/88.76 & 90.30/89.08 & \textbf{91.50}/90.61 & 90.11/89.61 & 89.74/88.71 & 91.40/90.45 & 91.22/90.49 \\ 
HALO-B (Ours, Ref.) & 91.59/90.58 & 92.24/91.40 & 93.25/92.39 & 93.62/92.91 & 93.25/92.43 & 93.35/92.65 & 93.53/92.83 & \textbf{93.99}/93.24 \\ 
HALO-L (Ours, Ref.) & 91.77/90.85 & 93.35/92.73 & 93.53/93.01 & 94.09/93.25 & 93.81/93.10 & 93.99/93.33 & 94.36/93.44 & \textbf{94.55}/93.81 \\ 
\bottomrule 
\end{tabular} 
} 
\end{table*}

\section{Real-World Scene-Level Validation on ScanNet}
\label{sec:supp_scannet_validation}
COGS-Bench is designed primarily for controlled object-level attribute diagnosis. However, scene-level perception introduces additional factors, including background context, occlusion, clutter, and varying scene geometry. We therefore conduct a proof-of-concept experiment to test whether the two main attribute-usage observations also transfer to real-world scene-level 3DGS perception. 

\textbf{Dataset and task.} We use the ScanNet 3DGS data provided by SceneSplat~\cite{li2025scenesplat} for training and evaluation. The SceneSplat release also provides semantic labels aligned with its Gaussian primitives, which we use directly as supervision. We consider the standard 20-class ScanNet semantic segmentation task. To keep the experiment computationally tractable while following an established protocol, training uses the official ScanNet Data-Efficient 10\% Limited-Reconstruction (LR) split~\cite{hou2021exploring}. No scenes outside the official 10\% training subset are used for model training. Evaluation is performed on the complete validation split. 

\textbf{Scene preprocessing and sampling.} We use only the native Gaussian attributes (P/O/S/Q/C). Unlike object classification, scene-level 3DGS samples contain a variable and potentially very large number of Gaussian primitives. We partition each training scene into $6\,\mathrm{m} \times 6\,\mathrm{m}$ spatial chunks with a stride of $3\,\mathrm{m}$. Within each chunk, we first apply grid sampling with a grid size of $0.02\,\mathrm{m}$. We then randomly select a center and perform spherical sampling to obtain 16,384 Gaussian primitives, enabling mini-batch training while retaining substantially denser scene geometry than in object classification. Each chunk is sampled 16 times per epoch, such that repeated random spherical sampling provides broad spatial coverage of the training chunks. All scene-level models are trained for 100 epochs with batch size 8. The learning rate follows cosine annealing from $1\times10^{-3}$ to $1\times10^{-5}$. During validation, the complete scene is grid-sampled at $0.02\,\mathrm{m}$ and all sampled primitives are processed by the network. Predictions are then propagated to all evaluation points using nearest-neighbor assignment, allowing metrics to be computed over the complete validation scene rather than only the grid-sampled inputs. 

\textbf{Dense prediction adaptation of HALO.} HALO was originally introduced as an object-classification reference model and progressively downsamples Gaussian primitives before global pooling. For semantic segmentation, we retain its hierarchical encoder but replace the classification-only output with a lightweight dense prediction path. At each hierarchy level, an MLP projects the per-primitive features. These features are propagated by nearest-neighbor feature propagation to the first hierarchy, allowing every primitive to collect representations from multiple semantic levels. Multi-level features are then aggregated to predict a semantic label using a simple classifier for each input primitive. We use [4, 4, 8, 4] HA blocks with [64, 128, 256, 512] feature channels at the four hierarchy levels to accommodate the substantially denser scene input. The shared-processing and attribute-specific variants use the same scene preprocessing, hierarchy, decoder, optimization, and POSQC input; they differ only in how heterogeneous Gaussian attributes are processed before fusion. 

\textbf{Results.} \cref{tab:supp_scannet_semseg} evaluates two complementary questions. 
First, we test whether the early-fusion degradation identified by COGS-Bench persists in real-world scene-level perception by repeating the complete eight-input diagnostic protocol with PTv1. Full POSQC obtains 48.45\% mIoU and 80.95\% OA. Importantly, three attribute subsets outperform POSQC on both metrics: PO reaches 49.67\% mIoU / 81.14\% OA, PSQC reaches 49.34\% / 81.23\%, and POSQ achieves the best mIoU of 50.37\% / 81.08\%. In particular, POSQ improves over full POSQC by 1.92 mIoU points. However, this is not a general preference for using fewer attributes: PSQ, POC, PC, and P all perform below POSQC in mIoU. The full ablation therefore shows that the contribution of heterogeneous Gaussian attributes under early fusion is composition-dependent rather than monotonically increasing with the number of available attributes. This subset-over-full behavior is consistent with the diagnostic observations on COGS and capture-derived MACGS, and demonstrates that the phenomenon also persists in real-world scene-level 3DGS perception. 
Second, we test whether attribute-specific processing can mitigate this degradation. Using the identical full POSQC input, HALO with shared attribute processing obtains 43.81\% mIoU and 79.16\% OA, whereas attribute-specific processing reaches 49.83\% mIoU and 81.96\% OA, corresponding to gains of 6.02 mIoU and 2.80 OA points. The substantial improvement indicates that preserving attribute-specific evidence before fusion remains beneficial under real-world scene clutter, background context, and occlusion. Together, the PTv1 ablation and the controlled HALO comparison provide complementary evidence that (i) homogeneous early fusion does not consistently exploit all available Gaussian attributes, and (ii) role-preserving attribute processing can alleviate this issue beyond clean object-level classification.

\begin{table}[t] 
\centering 
\caption{ Real-world scene-level proof-of-concept on the standard ScanNet 20-class semantic segmentation task. Training uses the official ScanNet Data-Efficient 10\% split and evaluation uses the complete validation set. } 
\label{tab:supp_scannet_semseg} 
\setlength{\tabcolsep}{4pt} 
\resizebox{\linewidth}{!}{ 
\begin{tabular}{llcc} 
\toprule 
Method & Variant/Input & mIoU (\% $\uparrow$) & OA (\% $\uparrow$) \\ 
\midrule 
PTv1~\cite{zhao2021point} & P     & 45.64 & 78.84 \\
PTv1~\cite{zhao2021point} & PO    & 49.67 & 81.14 \\
PTv1~\cite{zhao2021point} & PC    & 47.81 & 80.50 \\
PTv1~\cite{zhao2021point} & POC   & 46.70 & 80.49 \\
PTv1~\cite{zhao2021point} & PSQ   & 48.05 & 79.71 \\
PTv1~\cite{zhao2021point} & POSQ  & \textbf{50.37} & 81.08 \\
PTv1~\cite{zhao2021point} & PSQC  & 49.34 & \textbf{81.23} \\
PTv1~\cite{zhao2021point} & POSQC & 48.45 & 80.95 \\
\midrule 
HALO & Shared POSQC & 43.81 & 79.16 \\ 
HALO & Attribute-specific POSQC & \textbf{49.83} & \textbf{81.96} \\ \bottomrule 
\end{tabular} 
} 
\end{table}

\section{Inference Efficiency}
\label{sec:supp_efficiency}
The main-paper classification benchmark reports parameter count, peak GPU memory during training, and total training-and-evaluation time. To provide a more complete characterization of the accuracy--efficiency trade-off, we further measure arithmetic complexity, actual GPU inference latency, inference throughput, and peak inference memory under the same unified classification setting.

\textbf{Measurement protocol.}
All models are evaluated using the full POSQC input with 1,024 Gaussian primitives per object, consistent with the main classification benchmark. Measurements are performed in FP32 on the same NVIDIA RTX~4090 GPU with models in evaluation mode and gradient computation disabled. Inputs are preloaded on the GPU; therefore, disk I/O, data loading, host-to-device transfer, and the common dataset-level preprocessing are excluded. In contrast, all model-specific operations executed inside the forward pass, including neighborhood construction, grouping, and hierarchical downsampling, are included in the measured latency and memory.

Arithmetic FLOPs are measured for a single object ($B=1$) using \texttt{fvcore}. Following the \texttt{fvcore} convention, one fused multiply-add is counted as one FLOP. Since the default counter does not support several point-cloud-specific operators, including some KNN queries, farthest-point sampling, top-$k$ operations, indexing, and custom CUDA kernels, we report these values as \emph{counted FLOPs} rather than a complete hardware-cost estimate. Unsupported operations are excluded from the FLOP totals but remain fully included in the measured inference latency and memory.

For latency, each model is first executed for 100 warm-up iterations. We then measure 1,000 forward passes using CUDA events and report the mean GPU execution time per complete batch. Measurements are conducted for both batch size 1 (B1) and batch size 32 (B32). B32 throughput is derived from the measured batch latency as $\frac{32}{\mathit{latency}}$ and is reported in objects per second. Peak inference memory is measured with \texttt{torch.cuda.max\_memory\_allocated()} during a forward pass after resetting the CUDA peak-memory statistics. The reported value therefore includes the resident model and input together with the output and temporary tensors allocated during inference.

\cref{tab:supp_inference_efficiency} summarizes the results. PointNet provides the lowest latency and highest B32 throughput, but at substantially lower recognition accuracy. HALO-B reaches 94.50\% OA with only 2.68M parameters and 0.18G counted FLOPs. Its B1 inference latency is 5.70ms, while its B32 batch latency is only 5.81ms, corresponding to a throughput of 5,504 objects/s. This is the second-highest B32 throughput among the evaluated methods while retaining substantially higher accuracy than the lightweight baselines. HALO-B also requires only 30.58MiB peak allocated memory at B1 and 404.39MiB at B32.

The results further illustrate why arithmetic FLOPs alone are insufficient for characterizing point-cloud inference efficiency. For example, PTv1 has the lowest counted FLOPs (0.10G) but requires 10.48ms for B1 inference. Its default \texttt{fvcore} trace does not account for several operations, including farthest-point sampling, KNN queries, and associated neighborhood operations, all of which contribute to the measured execution time. We therefore use the measured latency and peak memory together with counted FLOPs when comparing computational efficiency.

HALO-L increases the model capacity from 2.68M to 10.46M parameters and the counted FLOPs from 0.18G to 1.51G, while improving OA by only 0.20 percentage points (94.50\% to 94.70\%). Its B32 throughput also decreases from 5,504 to 3,932 objects/s. These results reinforce our positioning of HALO-B as the recommended accuracy--efficiency configuration, while HALO-L primarily serves as a capacity-scaling variant.

\begin{table*}[t]
\centering
\caption{
Inference efficiency under the unified 1,024-Gaussian classification protocol.
Counted FLOPs are obtained using \texttt{fvcore} with one fused multiply-add counted as one FLOP; unsupported point-cloud-specific operators are excluded from FLOPs but included in latency and memory measurements.
B1/B32 latency denotes the mean CUDA-event execution time of the complete batch after 100 warm-up and 1,000 measured forward passes.
B32 throughput is computed from the measured B32 batch latency.
Peak inference memory reports the maximum allocated GPU memory during one forward pass.
}
\label{tab:supp_inference_efficiency}
\setlength{\tabcolsep}{5.0pt}
\resizebox{\textwidth}{!}{
\begin{tabular}{lcccccc}
\toprule
Method
& OA (\% $\uparrow$)
& Params (M $\downarrow$)
& Counted FLOPs (G $\downarrow$)
& B1 / B32 Lat. (ms/batch $\downarrow$)
& B32 Throughput (obj/s $\uparrow$)
& B1 / B32 Peak Mem. (MiB $\downarrow$) \\
\midrule
PointNet~\cite{qi2017pointnet}
& 79.53
& 3.47
& 0.45
& 1.24 / 3.81
& 8394
& 34.24 / 303.31 \\

DGCNN~\cite{wang2019dynamic}
& 85.69
& 1.81
& 2.73
& 1.38 / 26.48
& 1208
& 76.11 / 1968.81 \\

PTv1~\cite{zhao2021point}
& 87.67
& 3.26
& 0.10
& 10.48 / 18.35
& 1744
& 27.48 / 237.56 \\

DeLA~\cite{chen2023decoupled}
& 88.48
& 5.34
& 0.73
& 3.49 / 7.92
& 4039
& 46.33 / 565.85 \\

PTv3~\cite{wu2024point}
& 89.11
& 38.87
& 6.32
& 30.95 / 71.62
& 447
& 271.25	 / 424.38 \\

Gaussian-MAE~\cite{ma2024shapesplat}
& 88.20
& 22.10
& 2.44
& 4.01 / 8.84
& 3619
& 104.68 / 484.68 \\

HALO-B (Ours, Ref.)
& 94.50
& 2.68
& 0.18
& 5.70 / 5.81
& 5504
& 30.58 / 404.39 \\

HALO-L (Ours, Ref.)
& 94.70
& 10.46
& 1.51
& 5.75 / 8.14
& 3932
& 66.49 / 502.91 \\
\bottomrule
\end{tabular}
}
\end{table*}

\section{Additional HALO Ablations}
\label{sec:supp_ablation}
Although the main contribution of this work is COGS and COGS-Bench, we include component ablations of HALO to validate the attribute-aware reference model used in the benchmark. 
\cref{tab:supp_ablation_ha} evaluates the role of hierarchical aggregation. Removing hierarchical aggregation and directly pooling multi-level features leads to 84.26\% OA. Adding cross-level fusion without neighborhood aggregation brings only a small improvement, reaching 85.07\% OA. Omitting downsampling retains more primitives and improves accuracy to 88.97\%, but increases memory consumption to 1885 MB. The complete HA module achieves 94.50\% OA with only 750 MB memory. These results show that aggregating neighborhood evidence before primitive decimation is important for both accuracy and efficiency.

\begin{table}[t]
\centering
\caption{
Ablation study of hierarchical aggregation on COGS-Bench. ``--'' indicates that the component is not applicable. $\circ$ denotes that downsampling is omitted.
}
\label{tab:supp_ablation_ha}
\setlength{\tabcolsep}{5pt}
\resizebox{\linewidth}{!}{
\begin{tabular}{cc|ccc}
\toprule
Cross-level Fusion & Neighborhood Agg. & OA (\% $\uparrow$) & mAcc (\% $\uparrow$) & Mem. (MB $\downarrow$) \\
\midrule
$\times$ & -- & 84.26 & 85.15 & 482 \\
\checkmark & $\times$ & 85.07 & 85.68 & 603 \\
\checkmark & $\circ$ & 88.97 & 90.03 & 1885 \\
\textbf{\checkmark} & \textbf{\checkmark} & \textbf{94.50} & \textbf{94.70} & \textbf{750} \\
\bottomrule
\end{tabular}
}
\end{table}

\section{Full Cross-Pipeline Classification Results}
\label{sec:supp_cross_pipeline}
For completeness, \cref{tab:supp_external_results} reports the full-POSQC classification comparison on capture-derived MACGS and CAD-derived ModelNet40GS. All models are trained and evaluated within each target dataset without COGS pre-training, following the unified classification protocol in the main paper.

\begin{table}[t]
\centering
\caption{Full-POSQC cross-pipeline classification results on MACGS and ModelNet40GS.}
\label{tab:supp_external_results}
\setlength{\tabcolsep}{4pt}
\resizebox{\linewidth}{!}{
\begin{tabular}{l|cc|cc}
\toprule
\multirow{2}{*}{Method}
& \multicolumn{2}{c|}{MACGS} & \multicolumn{2}{c}{ModelNet40GS} \\
\cmidrule(lr){2-3} \cmidrule(lr){4-5}
& OA (\% $\uparrow$) & mAcc (\% $\uparrow$) & OA (\% $\uparrow$) & mAcc (\% $\uparrow$) \\
\midrule
PointNet~\cite{qi2017pointnet} & 79.76 & 77.97 & 90.98 & 87.79 \\
DGCNN~\cite{wang2019dynamic} & 87.15 & 85.80 & 90.82 & 87.50 \\
PTv1~\cite{zhao2021point} & 88.26 & 87.54 & 91.06 & 88.17 \\
DeLA~\cite{chen2023decoupled} & 91.50 & 90.15 & 91.99 & 89.60 \\
PTv3~\cite{wu2024point} & 92.33 & 91.45 & 91.97 & 89.14 \\
Gaussian-MAE~\cite{ma2024shapesplat} & 91.22 & 90.49 & 92.88 & 90.52 \\
\midrule
\textbf{HALO-B (Ours, Ref.)} & \underline{93.99} & \underline{93.24} & \underline{93.24} & \underline{91.14} \\
\textbf{HALO-L (Ours, Ref.)} & \textbf{94.55} & \textbf{93.81} & \textbf{93.53} & \textbf{91.33} \\
\bottomrule
\end{tabular}
}
\end{table}

\section{Additional COGS Visualizations}
\label{sec:supp_visualization}
We provide additional multi-view visualizations of COGS samples. Each sample is rendered from multiple viewpoints to show object-level structure, appearance variation, and view consistency. These visualizations complement the dataset examples and quality analysis in the main paper.

\begin{figure*}[t]
\centering
\includegraphics[width=\linewidth]{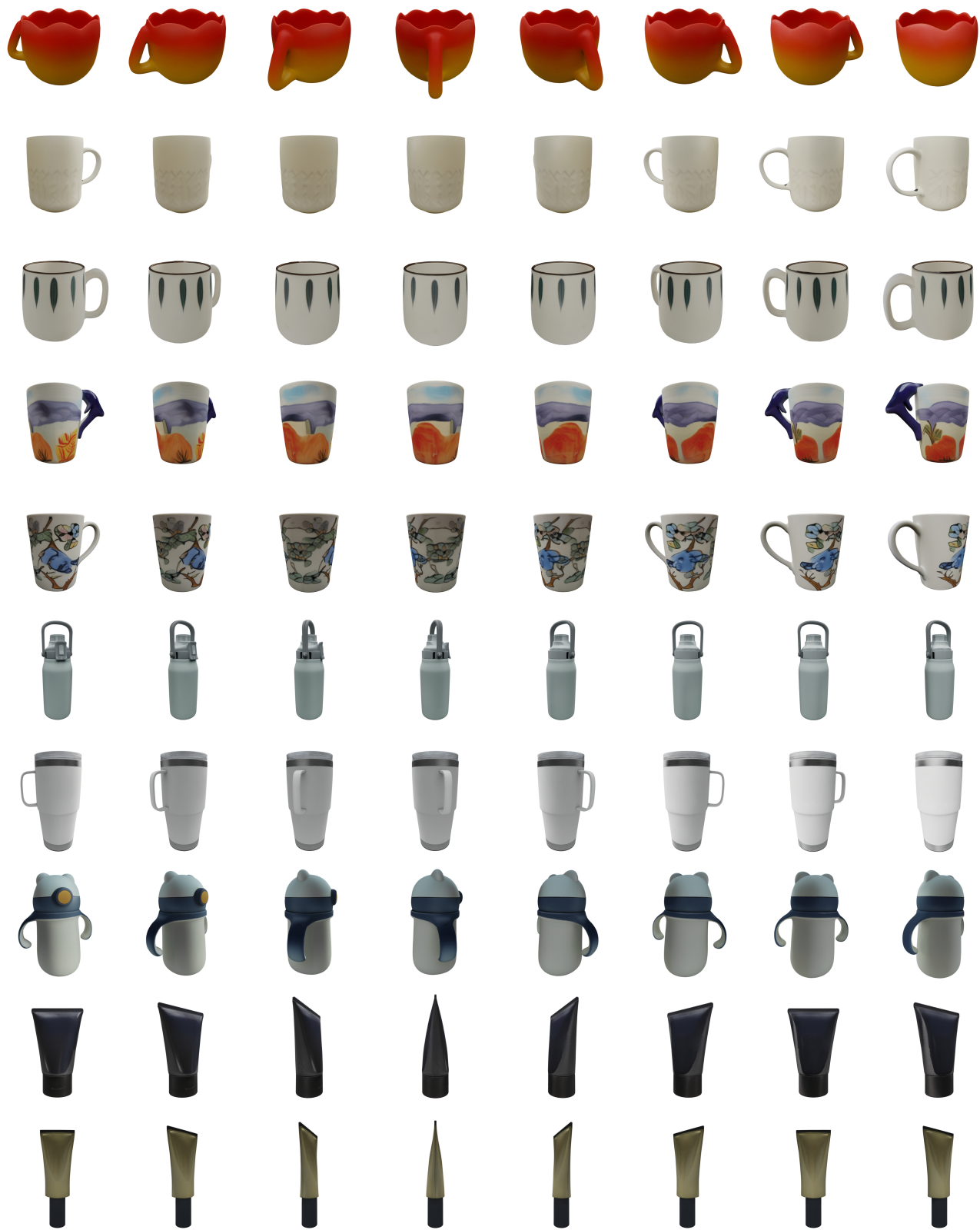}
\caption{
Additional multi-view visualizations of COGS samples. 
}
\label{fig:supp_cogs_samples_1}
\end{figure*}

\begin{figure*}[t]
\centering
\includegraphics[width=\linewidth]{images/dataset_samples_2.jpg}
\caption{
Additional multi-view visualizations of COGS samples.
}
\label{fig:supp_cogs_samples_2}
\end{figure*}

\begin{figure*}[t]
\centering
\includegraphics[width=\linewidth]{images/dataset_samples_3.jpg}
\caption{
Additional multi-view visualizations of COGS samples. 
}
\label{fig:supp_cogs_samples_3}
\end{figure*}

\begin{figure*}[t]
\centering
\includegraphics[width=\linewidth]{images/dataset_samples_4.jpg}
\caption{
Additional multi-view visualizations of COGS samples. 
}
\label{fig:supp_cogs_samples_4}
\end{figure*}

\begin{figure*}[t]
\centering
\includegraphics[width=\linewidth]{images/dataset_samples_5.jpg}
\caption{
Additional multi-view visualizations of COGS samples. 
}
\label{fig:supp_cogs_samples_5}
\end{figure*}

\begin{figure*}[t]
\centering
\includegraphics[width=\linewidth]{images/dataset_samples_6.jpg}
\caption{
Additional multi-view visualizations of COGS samples. 
}
\label{fig:supp_cogs_samples_6}
\end{figure*}

\end{document}